\documentclass{article}
\usepackage{longtable}
\usepackage{booktabs}
\usepackage{amsmath, amssymb, amsthm}
\usepackage{xcolor}
\usepackage{graphicx}
\usepackage{times}
\usepackage{epsfig}
\usepackage{graphicx}
\usepackage{amsmath}
\usepackage{amssymb}
\usepackage{multirow}
\usepackage{booktabs}
\usepackage{threeparttable}
\usepackage{subfig}
\usepackage{color}
\usepackage{cite}
\usepackage{eucal}
 \usepackage[backref]{hyperref} 
\begin{document}
	\title{G-DARTS-A: Groups of Channel Parallel Sampling with Attention}
	\author{Zhaowen Wang, Wei Zhang, Zhiming Wang\\
	University of Secience and Technology Beijing}

	\maketitle
	\begin{abstract}Differentiable Architecture Search (DARTS) provides a baseline for searching effective network architectures based gradient, but it is accompanied by huge computational overhead in searching and training network architecture. Recently, many novel works have improved DARTS. Particularly, Partially-Connected DARTS (PC-DARTS) proposed the partial channel sampling technique which achieved good results. In this work, we found that the backbone provided by DARTS is prone to overfitting. To mitigate this problem, we propose an approach named Group-DARTS with Attention (G-DARTS-A), using multiple groups of channels for searching. Inspired by the partially sampling strategy of PC-DARTS, we use groups channels to sample the super-network to perform a more efficient search while maintaining the relative integrity of the network information. In order to relieve the competition between channel groups and keep channel balance, we follow the attention mechanism in Squeeze-and-Excitation Network. Each group of channels shares defined weights thence they can provide different suggestion for searching. The searched architecture is more powerful and better adapted to different deployments. Specifically, by only using the attention module on DARTS we achieved an error rate of 2.82\%/16.36\% on CIFAR10/100 with 0.3GPU-days for search process on CIFAR10. Apply our G-DARTS-A to DARTS/PC-DARTS, an error rate of 2.57\%/2.61\% on CIFAR10 with 0.5/0.4 GPU-days is achieved.
	\end{abstract}
	\section{Introduction}\label{intro}
	Neural architecture search (NAS), as an emerging field of machine learning, has become one of the main methods to realize automatic machine learning, and has increasingly received widespread attention from the academic.\cite{elsken2018neural} The main technology of NAS is to construct various search spaces, design algorithms to search iteratively, train continuously and evaluate continuously to discover the optimal network architecture. After Google\cite{zoph2016neural} and MIT\cite{he2016deep} released NAS based on reinforcement learning in 2016, through improvements in search space\cite{zoph2018learning}, search strategies\cite{pham2018efficient}, and evaluation methods\cite{tan2019efficientnet} in recent years, NAS algorithms have gradually reduced the demand for GPUs and become more "civilized", enabling more people to have opportunities to provide new idea which promote the development of automated machine learning.
	Especially, DARTS\cite{hanxiao2019darts}, as one of the algorithms that rapidly improve the computational efficiency of NAS method, uses a neural network structure composed of stacked cells as the search space, and relax the Categorical selection of a specific operation to a softmax of all possible operations to make the search space continuous. In addition to the huge computational burden DARTS has to bear, the generalization ability of the searched network architecture also has space for improvement. The prior work named Partially-Connected DARTS (PC-DARTS)\cite{xu2019pc} proposed to sample a small part of super-network to reduce redundancy when exploring the network space. PC-DARTS improved the performance of the searched network architecture with only one group of channels, which indicated group of channels has good value for architecture search.
	In addition, Squeeze-and-Excitation Network (SENet)\cite{hu2018squeeze} that appeared in 2017 obtained the weight of different channels by learning the relationship between each channel, then multiplied the weight by the original feature map to acquire the final feature. We referred to the method of observing the connection between channels of attention mechanism and proposed our own parameters to balance the relationship between channels.
	In this paper, we present a novel approach named Group-DARTS with attention (G-DARTS-A) to reduce the overfitting and improve the stability in searching process. There are two primary contributions of our method, namely, Channel Group Parallel Sampling for Searching and the attention for balance between each group. As the key technique in this work, Channel Group Parallel Sampling has not been researched in NAS for searching more stable architecture with better performance. This point proposes a new idea to design super network for NAS. This kind of super network we built, combining Channel Group Parallel Sampling and weights for attention, can ensure the stability of the search process and prevent the searched architecture from overfitting.
	We increase the accuracy on cifar10 and cifar100 of PC-DARTS and DARTS. Moreover, these two components are easily transplanted to other search algorithms to improve search accuracy and generalization, e.g., these can boost the accuracy and speed of all algorithms based on DARTS search space. We only use the attention module of this method on DARTS and achieve an error rate of 2.82\%/16.36\% on CIFAR10/100 with 0.2GPU-days for search process on CIFAR10. Apply our G-DARTS-A to DARTS/PC-DARTS, an error rate of 2.57\%/2.64\% on CIFAR10 with 0.5/0.4 GPU-days is achieved.

	\section{Related work}\label{rl}
	Due to the increase in the amount of data and the improvement of computer computing power, people began to pursue to build more complex artificial neural networks. The existing excellent neural network structures built manually are time-consuming and error prone\cite{elsken2018neural} \cite{zoph2016neural} \cite{cai2019proxylessnas}. In order to realize automatic machine learning, neural architecture search as a process of automated architecture functional engineering enables the computer to automatically obtain the optimal network structure framework in the process of iterative search, continuous training and continuous evaluation.
	In 2016, Google published a neural architecture search based on reinforcement learning, which became the first well-known NAS work\cite{zoph2016neural}. It used recurrent neural network (RNN) trained through reinforcement learning to generate description of neural network model. After that, NASnet\cite{zoph2018learning} simplified searching the optimal convolutional architecture into finding the best units of CNN. With further improvement, AmoebaNet\cite{real2019regularized} generated an architecture based on genetic algorithms, surpassing manual design for the first time. The above algorithms rely on powerful hardware and bring huge computational overhead. In order to solve the computational problem, Google proposed an efficient neural architecture search (ENAS, Efficient NAS)\cite{pham2018efficient}, which used a controller to search for the best subgraph in a large computational graph to display the neural network architecture, and the subgraphs share parameters with each other. Then, ProxylessNAS\cite{cai2019proxylessnas} directly searched on the target data set instead of convention migration search strategies for reducing hardware requirements.
	In 2017, Senet\cite{hu2018squeeze} won the championship of the ImageNet 2017 competition classification task. SENet\cite{hu2018squeeze} uses the attention mechanism to focus on the relationship between channels, hoping the model can automatically learn the significance of different channel features. It learns the relationship between each channel to obtain the weight of different channels and multiplies the weight by the original feature map to obtain the final feature. This attention mechanism allows the model to pay more attention to the channel features with the most information, while suppressing those unimportant channel features.
	This work based on differentiable architecture search method, such as DARTS\cite{hanxiao2019darts} and PC-DARTS\cite{xu2019pc} and refers to SENet's\cite{hu2018squeeze} method of observing channel relationships. DARTS used the neural network structure stacked by computing units as the search space and reduced the search neural architecture to search for suitable units, which promoted a rapid increase in the computational efficiency of NAS method. It used bilevel optimization method for optimizing network weight and architecture parameter simultaneously. PC-DARTS\cite{xu2019pc} proposed channel sampling technology and edge normalization method, which shorten the search process of NAS and reduced the computational burden. This method only operated on a small part of the channels, then the remaining channels are merged with the operated channels after skipping the operating part. Structural parameters are optimized by randomly sampled channels in the iterative process.
	
	\section{Revisiting Gradient based Architecture Search}\label{nas}
	\subsection{Preliminaries of DARTS: continuous relaxation search space}\label{darts}
	DARTS is mentioned for the first time as the baseline of this work. Its search network is composed of a neural network formed by stacking L cells. Each cell can be regarded as a directed acyclic graph (DAG) connected by N nodes, $\{x_0,x_1,x_2,...,x_{N-1}\}$.
	Each pair of nodes $(i,j)$ is connected by edge $\mathrm{E}_{(i, j)}$ which is associated with different candidate structural operations, such as convolutional layer and pooling layer. We denote operations by $\mathcal{O}$, and the fixed function as elements in each operation is denoted by $o(\cdot)$. The core goal of DARTS is to make the search space continuous by using the architecture hyper-parameter $\alpha_o^{(i,j)}$ to quantify operation $o(x_i)$, so that the connection between edges $i$ and$j$ can be formulated as:
	$f_{i, j}(x_i) = \sum_{o\in\mathcal{O}_{i, j}}{\frac{\mathrm{exp}(\alpha_o^{(i,j)})}{\sum_{o'\in\mathcal{O}}\mathrm{exp}(\alpha_{o'}^{(i,j)})}o(x_i)}$.In addition, each cell consists of two input nodes, four intermediate nodes and one output node. The two input nodes are subject to the output of the first two cells respectively, and the output of this unit cell obtained by cascading all intermediate nodes.
	
	\subsection{Preliminaries of PC-DARTS: partial channel connections}
	In order to reduce the burden of memory and computer, PC-DARTS proposed a partial sampling method. Different from selecting all channels for candidate operation in each feature map, only 1/K channels are randomly sampled, followed by operation selection, while other shielded channels have no operation selection is performed. Because only part of the channel is operated, the burden of computer storage will be reduced by nearly K times, which makes it possible to use a batch size of K times larger than before in the search process. Not only does it speed up the network search, it also stabilizes the search process, making it possible to directly search for large-scale dataset.
	\section{Method}\label{mt}
	\subsection{Channel Group Parallel Sampling for Searching}
	
	The drawback of DARTS is simply training architecture parameters $\alpha$ as normal network weights, which increases the risk for overfitting in the search process. In addition, PC-DARTS with partial channel group connection can achieve good accuracy, which shows that the channel group has great value for searching architecture. In this work, we recommend using multiple channel group to design a super network, which named Channel Group Parallel Sampling method depicted in figure 1. In each cell, we set M edges on each pair of nodes, and each edge on the same connection shares $\beta$ weight. the method divides the feature maps into M groups that are seeded to the corresponding edges. Finally, the number of channels on each edge and M can be adjusted flexibly to adapt to various application scenarios.\\
	\begin{figure}[t]
	\centering
	\includegraphics[width=0.9\linewidth]{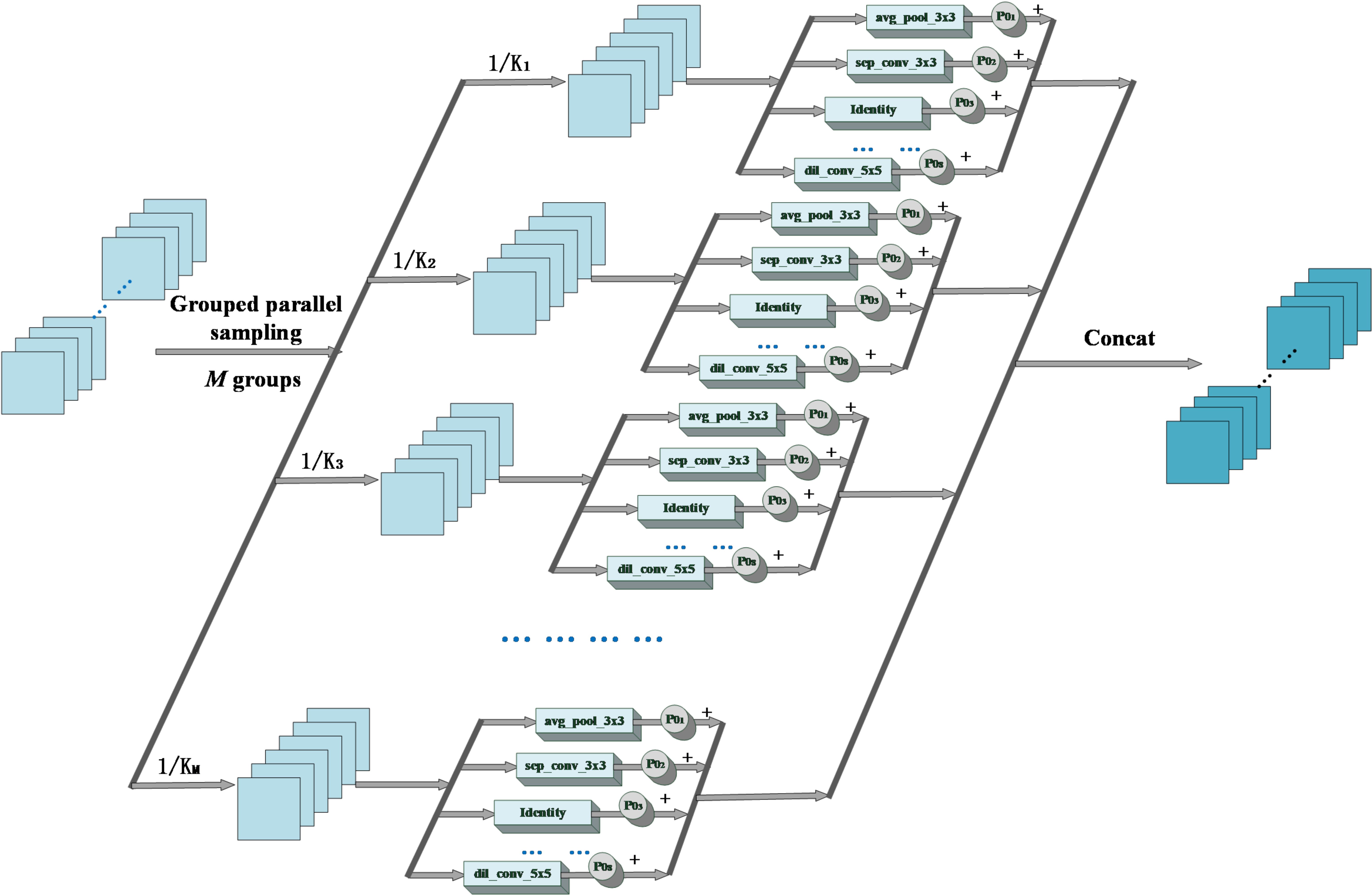}
	\caption{Illustration of Channel Group Parallel Sampling (best viewed in color), In the sprite of PC-DARTS, we show the propagation process of information between nodes. $\mathcal P_{0i}$ denote operations (denoted by O) on each edge (between node i and j).}
	\label{motivation}
	\end{figure}	
	This brings twofold benefits: First, multiple groups can bring more and different information to search process, so that the super-network has different optimization directions in weight updating. Additionally, the searched architecture can adapt to different groups of channels, thus network structure obtained in this way has stronger stability and generalization ability, which can effectively avoid overfitting of structural parameters. Therefore, the architecture searched by our G-DARTS-A also perform well on other data sets that are not used in the search process.

	\subsection{Attention for Balance between Each Channel Group}
	When it comes to Channel Group Parallel Sampling, it has advantages and disadvantages. The positive side was mentioned in the previous section. On the downside, first, multi-channel group greatly reduce the number of parameters in super network, which causes the decline of the super network's fitting ability. Moreover, different group of channels have unequal importance for searching, which may lead to vicious competition between each group of channels.\\ 
	\begin{figure}[t]
	\centering
	\includegraphics[width=0.9\linewidth]{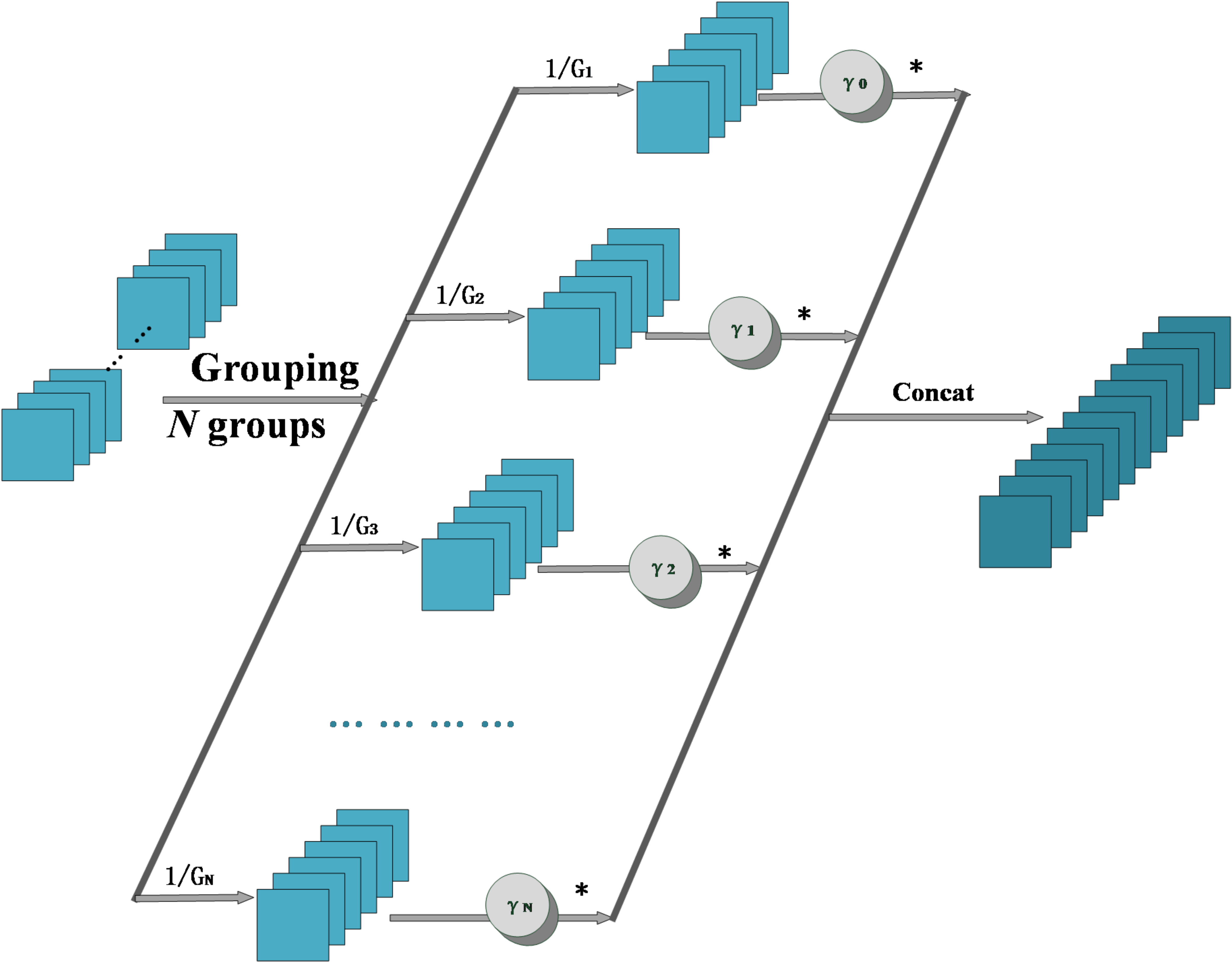}
	\caption{Representation of our weights.$\mathcal \gamma_0 $ to $\mathcal \gamma_N $represent the weights for Attention multiplied on the groups of channels.}
	\label{motivation}
	\end{figure}
	This disadvantage implies that the network has not learned deeply about the relationship between channel group. Simultaneously, we are inspired by the previous work of SEnet \cite{hu2018squeeze}, to alleviate these problem, we set weights that can be gradient as the channel group weights on each edge$(i,j)$, denoted by $\gamma$ showed as figure 2. Specifically, the weights of each edge in different cell is independent, which gives a large space for optimizing weights in search process. The bigger space for searching imply stronger ability to fit. Moreover, the weights can give more important groups of channels bigger weights than other groups of channels, which can keep a balance between each group of channels for searching a stable architecture with better performance.

	The two main ideas we propose can be combined into Channel Group Parallel Sampling with Attention, which be expressed by Equation 1.
	\begin{align*}
  	& {{f}_{i,j}}\left( {{x}_{ij}} \right)={{\gamma }_{1}}\sum\limits_{o\in O}{\frac{\exp \left\{ \alpha _{i,j}^{o} \right\}}{\sum\limits_{{{o}^{\prime }}\in O}{\exp }\left\{ \alpha _{i,j}^{{{o}^{\prime }}} \right\}}}o\left( {{x}_{i1}}\  \right)+{{\gamma }_{2}}\sum\limits_{o\in O}{\frac{\exp \left\{ \alpha _{i,j}^{o} \right\}}{\sum\limits_{{{o}^{\prime }}\in O}{\exp }\left\{ \alpha _{i,j}^{{{o}^{\prime }}} \right\}}}o\left( {{x}_{i2}}\  \right) \\
 &+{{\gamma }_{3}}\sum\limits_{o\in O}{\frac{\exp \left\{ \alpha _{i,j}^{o} \right\}}{\sum\limits_{{{o}^{\prime }}\in O}{\exp }\left\{ \alpha _{i,j}^{{{o}^{\prime }}} \right\}}}o\left( {{x}_{i3}}\  \right)+{{\gamma }_{4}}\sum\limits_{o\in O}{\frac{\exp \left\{ \alpha _{i,j}^{o} \right\}}{\sum\limits_{{{o}^{\prime }}\in O}{\exp }\left\{ \alpha _{i,j}^{{{o}^{\prime }}} \right\}}}o\left( {{x}_{i4}}\  \right) \\
 & +\ldots +{{\gamma }_{M}}\sum\limits_{o\in O}{\frac{\exp \left\{ \alpha _{i,j}^{o} \right\}}{\sum\limits_{{{o}^{\prime }}\in O}{\exp }\left\{ \alpha _{i,j}^{{{o}^{\prime }}} \right\}}}o\left( {{x}_{iM}}\  \right) \tag{Equation 1}
	\end{align*}
	
	For Equation 1, We show the operation of the channel group between nodes i and j.$\mathcal x_{i1-M} $represents different channel groups of the operating channel,$ \sum\limits_{o\in O}{\frac{\exp \left\{ \alpha _{i,j}^{o} \right\}}{\sum\limits_{{{o}^{\prime }}\in O}{\exp }\left\{ \alpha _{i,j}^{{{o}^{\prime }}} \right\}}}$ represents the operation to be performed by the channel group, which gives different weights $\mathcal \alpha_{i,j}^{o^\prime}$ to the operation (convolutional layer, pooling layer, etc.). The final channel group operation result between the two nodes (i to j) is expressed as $\mathcal f_{i,j}\left(x_{ij}\right)$. In addition, the Attention can be divided into different groups for weighting multiplication, shown on Equation 1 which contains the Attention and Channel Group Parallel Sampling with same grouping. A more detailed presentation is shown in Figure 3.

	\begin{figure*}[t]
	\centering
	\includegraphics[width=0.9\linewidth]{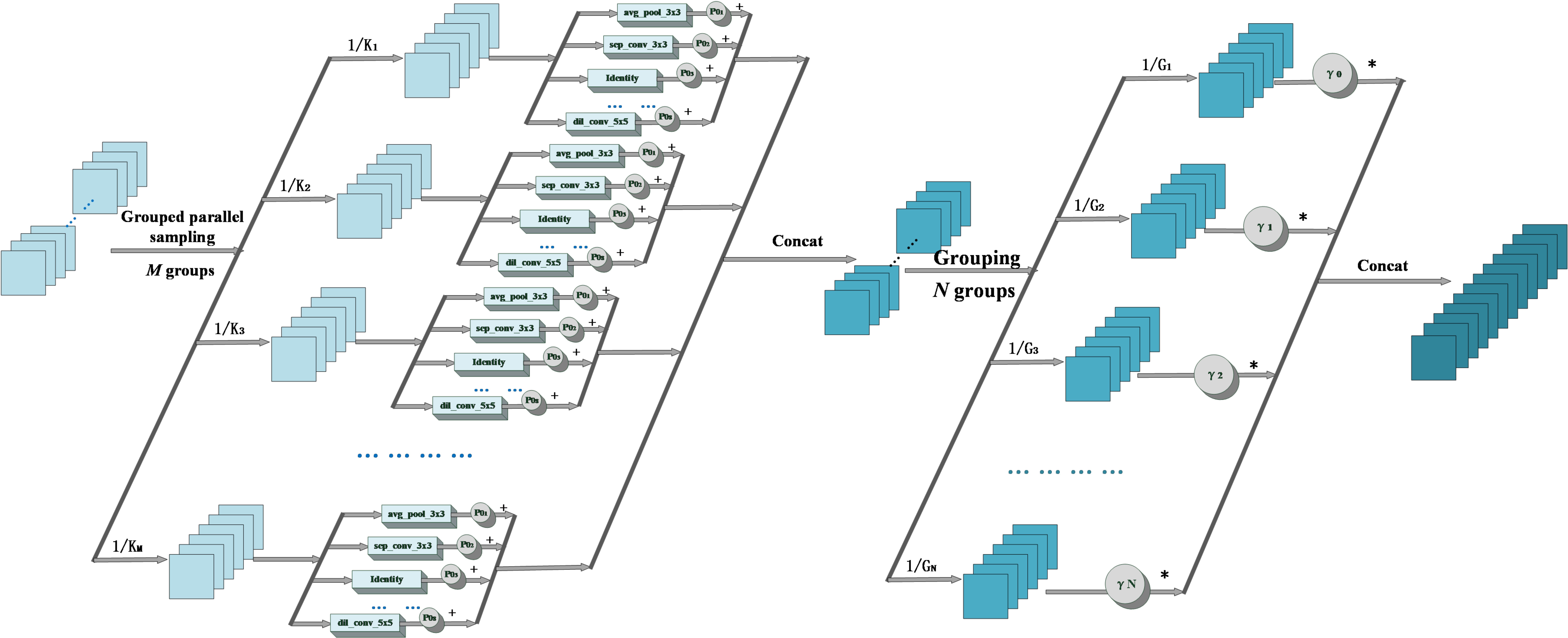}
	\caption{Illustration of Channel Group Parallel Sampling with Attention (best viewed in color) After M groups of parallel sampling, the feature map can be divided into N groups and multiplied by the attention weight $\mathcal \gamma_N $.}
	\label{motivation}
	\end{figure*}
	\subsection{Grouped convolution improves the speed}
	Due to the characteristics of GPU and Pytorch\cite{paszkeautomatic} calculations, multi-channel group parallelism will cause the search speed to slow down and grouped sampling of features mainly changes the characteristics of convolution operations. Inspired by this, we use the grouped convolution inside Pytorch for approximate replacement. Grouped convolution groups the input feature maps, then each group is convolved separately. Assuming that $c$ represents the number of channels of the input feature map, and it is set to be divided into g groups, the number of input feature maps in each group becomes $c/g$, and the size of the convolution kernel is correspondingly reduced to the original $c/g$ times. In this way, memory can be reduced while achieving a speed advantage, so that the algorithm achieves a similar effect with faster speed, thus we replaced all convolution operations in DARTS with grouped convolution. This improvement achieves the improvement of experimental accuracy without introducing any additional overhead.
	
	\section{Experiments}\label{ex}
	\subsection{Dataset}
	We conduct experiments on CIFAR10 and CIFAR100 datasets. Architecture search is performed on CIFAR10 and the searched architectures are evaluated on CIFAR10 and CIFAR100. CIFAR-10 contains $60$K RGB images which are equally divided into 10 categories. Each image has a spatial resolution of $32*32$. Among them, there are $50$K and $10$K images for training and testing. In other words, CIFAR10 has $5$ training batches and $1$ test batch, and each batch has $10$K images. In each training batch, the number of images of different types may not be the same but still accurately contains 5000 images of each class. The only difference between CIFAR100 and CIFAR10 is that CIFAR100 divides $60$K images into $100$ categories on average.
	\subsection{Architecture Search on CIFAR10}\label{archi-se}
	\subsubsection{Implementation Details}
	We implement the core methods of G-DARTS-A on DARTS and PC-DARTS. In search process, training set are equally distributed into two parts for respectively tuning network weights and architecture parameters.
	DARTS-with-Attention for balance between each group of channels, we followed its defined search space which contains $8$ operations, i.e., $3*3$ average-pooling, $3*3$ max-pooling, skip-connect, $3*3$ and $5*5$ separable convolution, $3*3$ and $5*5$ dilated separable convolution and zero. We divide the features on each edge into four groups of $1/8$, $1/8$, $1/4$, $1/2$ according to the channel, and multiply them by four gradient-optimized parameters with different initial values: $2.4$, $2.4$, $3.2$ and $3.0$ correspond to four groups of channels respectively. After that,on G-DARTS-A we changed all the convolution in the operation to grouped convolution with the number of groups being $4$.\\
	We use DARTS (first order) to train a small network consisting of $8$ cells, which lasts $50$ epochs, the batch size is $32$, and the initial number of channels is $16$. We use momentum SGD\cite{polyak1992acceleration} to optimize the weight $w$, initial learning rate $\eta_w= 0.025$ (anneal it to zero according to the cosine schedule without restarting), momentum $0.9$ and weight decay $3\times 10^{-4}$, while we use Adam\cite{kingma2014adam} as the optimizer for $\alpha$ with initial learning rate$\eta_\alpha= 3\times 10^{-4}$, momentum $\beta = (0.5, 0.999) $and weight decay$ 0.001$, and cutout\cite{devries2017improved} status is True.
	
	\subsubsection{Search Results}
	The architecture found from G-PC-DARTS-A on CIFAR10 by this method tends to retain deeper connections than the architecture found in previous work, as shown in Figure4(a) and Figure4(b). The cells found from DARTS-with-Attention are shown in Figure4(c) and Figure4(d). Besides, Figure4(e) and Figure4(f) represent the deep network architecture obtained from G-DARTS-A that uses grouped convolution instead of channel groups for sampling and accompanied by attention. The more layers of the network, the higher the possibility of its classification accuracy. Besides, we use 1080TiGPU to search architecture, G-PC-DARTS-A spent 0.3GPU-days while DARTS-with-Attention used 0.4GPU-days in Pytorch.

	\begin{figure}[htbp]
	\centering
	\subfloat[ normal cells of G-PC-DARTS-A]{

		\label{fig:improved_subfig_a}
		\includegraphics [width=0.33\textwidth]{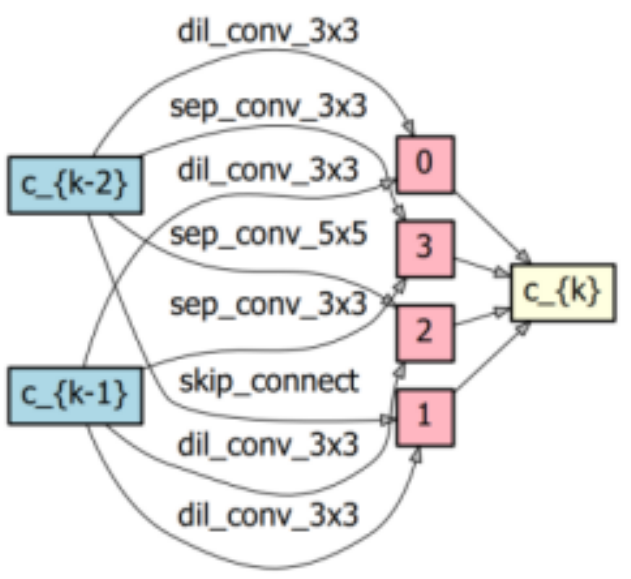}
		\hspace{10 pt}
	}
	\subfloat[ reduce cells of G-PC-DARTS-A]{

		\label{fig:improved_subfig_a}
		\includegraphics [width=0.66\textwidth]{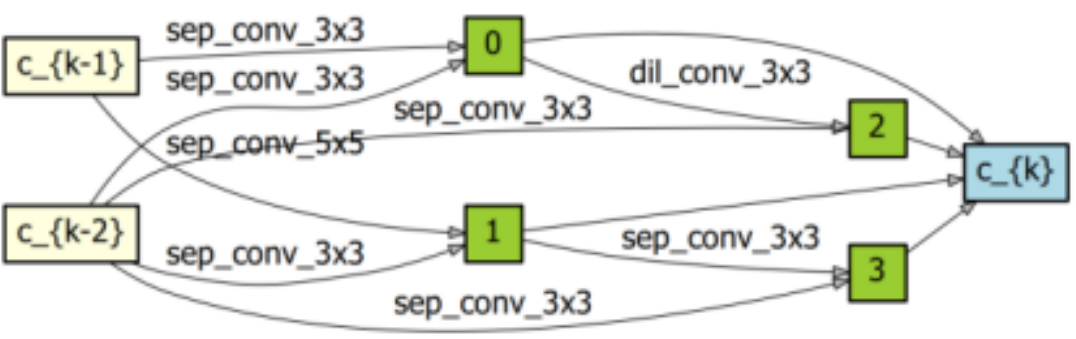}
		\hspace{10 pt}
	}
	\newline
	\subfloat[normal cells of DARTS-with-Attention]{

		\label{fig:improved_subfig_a}
		\includegraphics [width=0.5\textwidth]{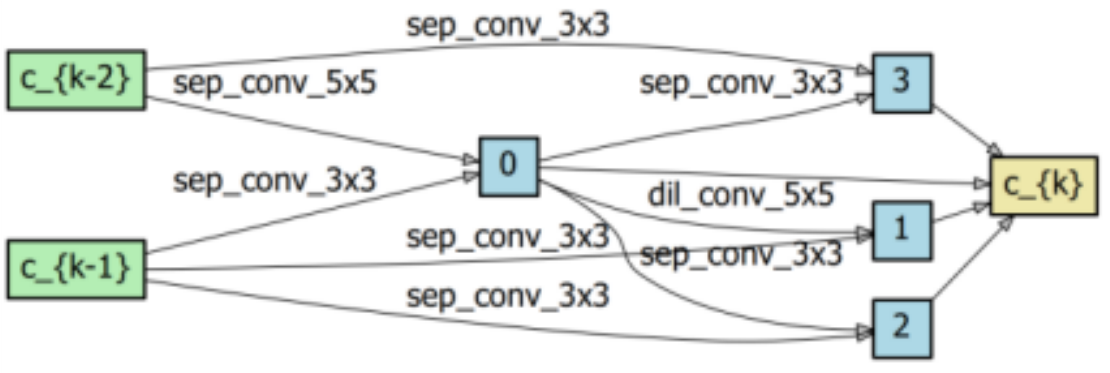}
		\hspace{10 pt}
	}
	\subfloat[ reduce cells of DARTS-with-Attention]{

		\label{fig:improved_subfig_a}
		\includegraphics [width=0.45\textwidth]{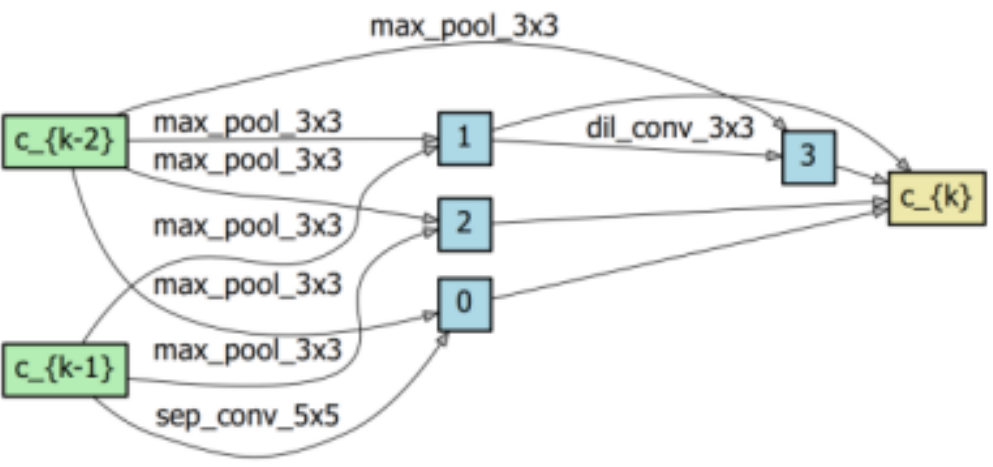}
		\hspace{10 pt}
	}
	\newline
	\subfloat[normal cells of G-DARTS-A]{

		\label{fig:improved_subfig_a}
		\includegraphics [width=0.8\textwidth]{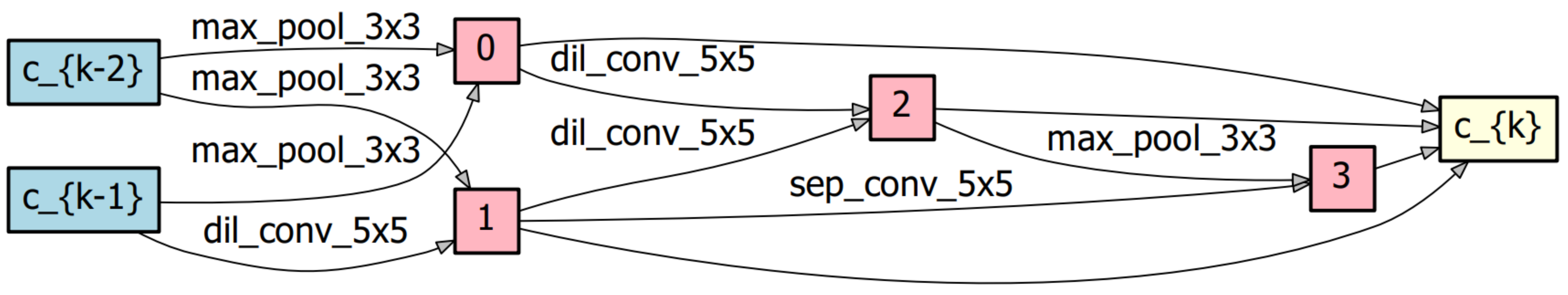}
		\hspace{10 pt}
	}
	\newline
	\subfloat[reduce cells of G-DARTS-A]{

		\label{fig:improved_subfig_a}
		\includegraphics [width=0.9\textwidth]{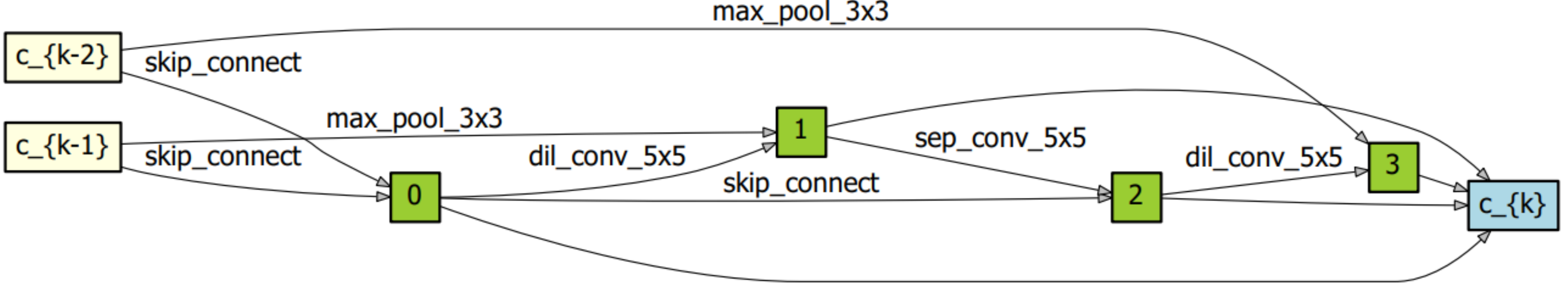}
		\hspace{10 pt}
	}
	\caption{Cells of G-PC-DARTS-A and DARTS-with-Attention found on CIFAR10} 
	\label{fig:subfig}
	\end{figure}
	\subsection{Architecture Evaluation on CIFAR10 and CIFAR100}
	For DARTS-with-Attention and G-PC-DARTS-A, we both train a large network of 20 cells for 600 epoch with batch size 96, and the initial number of channels is increased from 16 (search process) to 36. The remaining hyperparameters keep same as those used for searching architecture. Path dropout\cite{srivastava2014dropout} of probability 0.2 and auxiliary towers with weight 0.4. The size of searched architecture is 4.2M and 3.3M on DARTS-with-Attention and G-PC-DARTS-A respectively. Our searched architectures remain small size together with great performance. DARTS-with-Attention achieved 2.82\%/16.36\% rate of evaluation error on CIFAR10/CIFA100, and G-PC-DARTS-A provide 2.64\%/16.87\% error on CIFAR10/CIFAR100. The specific experimental data is shown in table1.

	\begin{table*}[t]
	%\begin{threeparttable}
	\caption{Comparison with state-of-the-art architectures on CIFAR10 and CIFAR100.}
	\begin{center}
	\resizebox{\textwidth}{!}{
	\begin{tabular}{lccccc}

	\hline\noalign{\smallskip}
	\textbf{\multirow{2}{*}{Architecture}} & \multicolumn{2}{c}{\textbf{Test Err. (\%})} & \textbf{Params} & \textbf{Search Cost} & \textbf{\multirow{2}{*}{Search Method}} \\
	\cmidrule(lr){2-3}
	&                            \textbf{C10} & \textbf{C100} & \textbf{(M)} & \textbf{(GPU-days)} &\\
	\noalign{\smallskip}\hline\noalign{\smallskip}
	DenseNet-BC~\cite{huang2017densely}                 & 3.46 & 17.18 & 25.6 & -    & manual \\
	\noalign{\smallskip}\hline\noalign{\smallskip}
	NASNet-A + cutout\cite{zoph2018learning}              & 2.65 & -     & 3.3  & 1\rm{,}800 & RL      \\
	AmoebaNet-B + cutout\cite{real2019regularized}        & 2.55 & -     & 2.8  & 3\rm{,}150 & evolution \\
	Hireachical Evolution\cite{liu2017hierarchical}       & 3.75 & -     & 15.7 & 300  & evolution \\
	PNAS\cite{liu2018progressive}                          & 3.41 & -     & 3.2  & 225  & SMBO \\
	ENAS + cutout\cite{pham2018efficient}                  & 2.89 & -     & 4.6  & 0.5  & RL \\
	NAONet-WS\cite{luo2018neural}                  & 3.53 & -     & 3.1  & 0.4  & NAO \\
	\noalign{\smallskip}\hline\noalign{\smallskip}
	
	DARTS (first order) + cutout\cite{hanxiao2019darts}       & 3.00 & 17.76$^\dagger$ & 3.3 & 1.5$^\ddagger$  & gradient-based \\
	DARTS (second order) + cutout\cite{hanxiao2019darts}       & 2.76 & 17.54$^\dagger$ & 3.3 & 4.0$^\ddagger$ & gradient-based \\
	SNAS + mild constraint + cutout\cite{xie2018snas}        & 2.98 & -     & 2.9  & 1.5  & gradient-based \\
	SNAS + moderate \\ constraint + cutout\cite{xie2018snas}    & 2.85 & -     & 2.8  & 1.5  & gradient-based \\
	BayesNAS  + cutout\cite{zhou2019bayesnas}  & 2.81±0.04 & -     & 3.4 & 0.2  & gradient-based \\
	ProxylessNAS + cutout\cite{cai2019proxylessnas}          & 2.08 & -     & 5.7  & 4.0  & gradient-based \\
	P-DARTS + cutout\cite{chen2019progressive}                     & 2.50 & 17.20 & 3.4  & 0.3 & gradient-based \\
	PC-DARTS + cutout\cite{xu2019pc}                     & 2.57 & - & 3.6  & 0.1 & gradient-based \\

	\noalign{\smallskip}\hline\noalign{\smallskip}
	
	DARTS (first order) with \\ Attention CIFAR10 + cutout         & 2.82 & 16.36     &4.2 & 0.4  & gradient-based \\
	G-DARTS (first order)-A \\ CIFAR10+cutout                    & 2.57 & 16.51 & 4.2  & 0.5 & gradient-based \\
	G-PC-DARTS with Attention \\ CIFAR10 + cutout                     & 2.61 & - & 2.8  & 0.3 & gradient-based \\
	\noalign{\smallskip}\hline
	\end{tabular}
	}
	\end{center}
	\label{tab_ev_cifar}
	\end{table*}

	\section{Conclusion}
	In this article, we propose a Channel Group Parallel Sampling technique with attention. The main method is to sample the feature map in multi-channel group to improve the computational efficiency while ensuring the integrity of the feature information. Attention weights contribute to balancing the relationship between each channel group, and we use it alone on DARTS and get an error rate of 2.82\%/16.36\% on CIFAR10/CIFAR100. Applying our complete technology to PC-DARTS, we get an error rate of 2.64\%/16.87\% on CIFAR10/CIFAR100.\\
	According to the experimental process and results of this work, the following four ideas are proposed. (i) The effective results of this work indicate that the value between architecture parameters and ordinary network weights is quite different, and a certain way is needed to distinguish them, especially by adding some parameters that are optimized only in the architecture search process to assist its optimization, such as our attention parameter and the edge normalization proposed by PC-DARTS. This work points out that different channel groups have independent and different values in the search process, which means we can take more flexible way to search for neural network structures. (iii) The results of this paper may imply excessive redundancy of model channel information. we can try to compress the model by reducing the number of channels together with adding an attention mechanism. (iv) These findings may exist in the generated training model. we can try to introduce multiple sets of parallelism and attention mechanisms to the final model. There is still various work worth exploring, and we will further supplement experiment to explore the value behind problem in more depth.

	\bibliographystyle{unsrt} 
	\bibliography{ref} 
\end{document}